\documentclass{article}
\usepackage[usenames, dvipsnames]{color}
\usepackage{amsmath,amsthm,amssymb}
\usepackage{graphicx}
\usepackage{subcaption}
\usepackage{blkarray}
\usepackage{comment}
\usepackage{soul}

\usepackage[final]{nips_2018}

\usepackage[utf8]{inputenc} 
\usepackage[T1]{fontenc}    
\usepackage{hyperref}       
\usepackage{url}            
\usepackage{booktabs}       
\usepackage{amsfonts}       
\usepackage{nicefrac}       
\usepackage{microtype}      

\title{Explainable Genetic Inheritance Pattern Prediction}
\author{
  Edmond Cunningham \\
  University of Michigan \\
  \texttt{edcunnin@umich.edu} \\
  \And
  Dana Schlegel \\
  University of Michigan \\
  \texttt{dschleg@med.umich.edu} \\
  \And
  Andrew DeOrio \\
  University of Michigan \\
  \texttt{awdeorio@umich.edu} \\
}

\begin{document}

\maketitle

\newcommand{\lemmaref}[1]{\ref{lemma:#1}}
\newcommand{\equationref}[2]{\ref{lemma:#1}.\ref{lemma:#1:#2}}

\newcommand{\mya}[2]{P(!(#1,#2)_y,{#1}_y,{#1}_x)}
\newcommand{\myb}[1]{P(\downarrow(#1)_y,{#1}_y|P(#1)_x)}
\newcommand{\myu}[1]{P(\uparrow(#1)_y,{#1}_y,{#1}_x)}
\newcommand{\myv}[2]{P(\downarrow(#1,#2)_y|{#1}_x)}

\newcommand{\downedgeand}[1]{\bigcap_{e \in \downedge{#1}}}
\newcommand{\downedgeandexcept}[2]{\bigcap_{e\prime \in \downedge{#1}\setminus #2}}
\newcommand{\downedgeprod}[1]{\prod_{e \in \downedge{#1}}}
\newcommand{\downedgeprodexcept}[2]{\prod_{ e\prime \in \downedge{#1}\setminus #2}}
\newcommand{\transition}[1]{P({#1}_y|{#1}_x)P({#1}_x|P(#1)_x)}
\newcommand{\siblingprod}[1]{\prod_{{#1}_s\in S(#1)}}
\newcommand{\parentprod}[1]{\prod_{{#1}_p\in P(#1)}}
\newcommand{\parentFprod}[1]{\prod_{{#1}_p\in P_F(#1)}}
\newcommand{\mateprod}[2]{\prod_{{#1}_m\in M(#1,#2)}}
\newcommand{\mateFprod}[2]{\prod_{{#1}_m\in M_F(#1,#2)}}
\newcommand{\childprod}[2]{\prod_{{#1}_c\in C(#1,#2)}}
\newcommand{\siblingand}[1]{\bigcap_{{#1}_s\in S(#1)}}
\newcommand{\parentand}[1]{\bigcap_{{#1}_p\in P(#1)}}
\newcommand{\mateand}[2]{\bigcap_{{#1}_m\in M(#1,#2)}}
\newcommand{\childand}[2]{\bigcap_{{#1}_c\in C(#1,#2)}}

\newcommand{\upedge}[1]{\wedge(#1)}
\newcommand{\downedge}[1]{\vee(#1)}
\newcommand{\upfrom}[1]{\uparrow(#1)}
\newcommand{\downfrom}[1]{\downarrow(#1)}
\newcommand{\downfromedge}[2]{\downarrow(#1,#2)}
\newcommand{\outfromedge}[2]{!(#1,#2)}

\newcommand{\setand}{\bigcap}
\newcommand{\setor}{+}
\newcommand{\bigCI}{\mathrel{\text{\scalebox{1.77}{$\perp\mkern-10mu\perp$}}}}
\newcommand{\doOp}[1]{\{#1\}}

\newcommand{\downedgeciexcept}[2]{\bigCI_{e\prime \in \downedge{#1}\setminus #2}}

\newcommand{\FIXME}[1]{\textcolor{red}{FIXME: #1}}
\newcommand{\oldfix}[1]{\textcolor{green}{#1}}
\newcommand{\newfix}[1]{\textcolor{black}{#1}}

\begin{abstract}
  Diagnosing an inherited disease often requires identifying the pattern of inheritance in a patient's family.  We represent family trees with genetic patterns of inheritance using hypergraphs and latent state space models to provide explainable inheritance pattern predictions.  Our approach allows for exact causal inference over a patient's possible genotypes given their relatives' phenotypes.  By design, inference can be examined at a low level to provide explainable predictions.  \newfix{Furthermore, we make use of human intuition by providing a method to assign hypothetical evidence to any inherited gene alleles.}  Our analysis supports the application of latent state space models to improve patient care in cases of rare inherited diseases where access to genetic specialists is limited.
\end{abstract}

\section{Introduction}
Genetic specialists rely heavily on a patient's family history to determine the inheritance pattern of a disease.  Using the inheritance pattern, specialists can narrow down the set of possible diagnoses.  Rather than mathematically calculating the most likely pattern of inheritance, the specialist uses knowledge and experience to identify high-level pedigree features in order to make a prediction.

Mendel's laws of inheritance describe how genotypes, a person's true genetic makeup, are transmitted from parents to children, and how they are expressed as phenotypes, the observed characteristics of their genetic condition.  The inheritance patterns considered in this paper are autosomal dominant (AD), autosomal recessive (AR), and X-linked recessive (XL).  Autosomal and X-linked recessive diseases are transmitted through mutated alleles of genes on non-sex chromosomes (autosomes) or on X chromosomes, respectively.  Dominant diseases require only a single mutated allele to be expressed, while recessive diseases require both alleles of the gene to be mutated.

Mendel's laws are not precisely followed during real-life reproduction.  \textit{De novo} mutations randomly introduce mutated alleles into a particular egg or sperm, and incomplete penetrance allows some individuals who have a mutated allele to escape symptoms of the condition.  Furthermore, some affected individuals may never receive a clinical diagnosis.  Although a person’s Mendelian disease status is determined by their genetic makeup, only their phenotypes are represented on a pedigree.

Pedigrees describe a patient's family history of disease (Figure \ref{fig:pedigree_example}).  Circles, squares and diamonds denote females, males and unknown sex.  Shaded nodes indicate a diagnosis of the familial disease.

\subsection{Contributions}
We propose an explainable prediction model that predicts a pattern of inheritance given a patient family history.  \newfix{To the best our knowledge, it is the first to apply latent state space algorithms to the inheritance pattern prediction problem.  Our implementation provides smoothed probabilities over genotypes per person and per family, and allows genetic specialists to evaluate hypotheses about who in the family tree might be a carrier in order to aid them make a grounded final prediction.} Our code is available at \cite{Cunningham2018}.

\section{Overview}
Our method uses latent state space models over hypergraphs to represent pedigrees.  In this section, we describe the dataset, why our model is a good choice for this problem, how it is possible to incorporate human intuition through \newfix{hypothetical evidence} and some ways to explain the outcome of a prediction.

\subsection{The Dataset}
We analyzed hand-drawn pedigrees that were written on paper during patient interviews.  During the interviews, a genetic specialist drew a pedigree based on the patient's familial history of disease.  A series of blood tests then confirmed the disease.  The pedigrees in our dataset were labeled with the pattern corresponding to the diagnosed disease.  We used digitized pedigrees that were collected by \cite{Schlegel17}.  An issue with pedigrees is that it is impossible to distinguish missing data from unaffected individuals -- both are denoted as unshaded nodes.  In addition, rates of incomplete penetrance and \textit{de novo} mutations are unknown for the rare diseases we examined.  Thus, generating synthetic data was unrealistic.

\subsection{Hypergraphs}
\begin{figure}
  \centering
  \begin{subfigure}{0.3\linewidth}
    \includegraphics[height=1in]{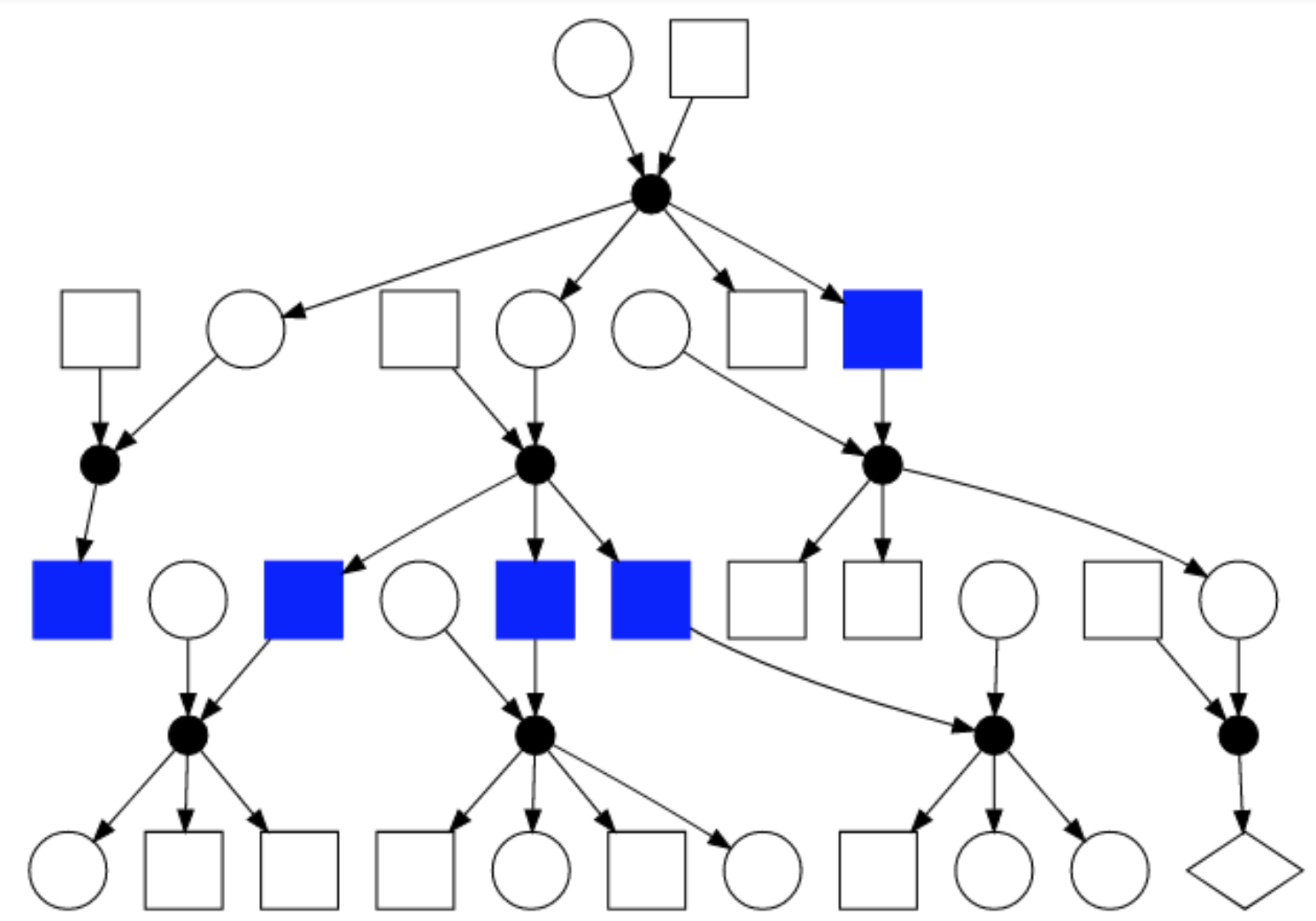}
    \caption{Example pedigree}
    \label{fig:pedigree_example}
  \end{subfigure}
  \begin{subfigure}{0.3\linewidth}
    \includegraphics[height=1in]{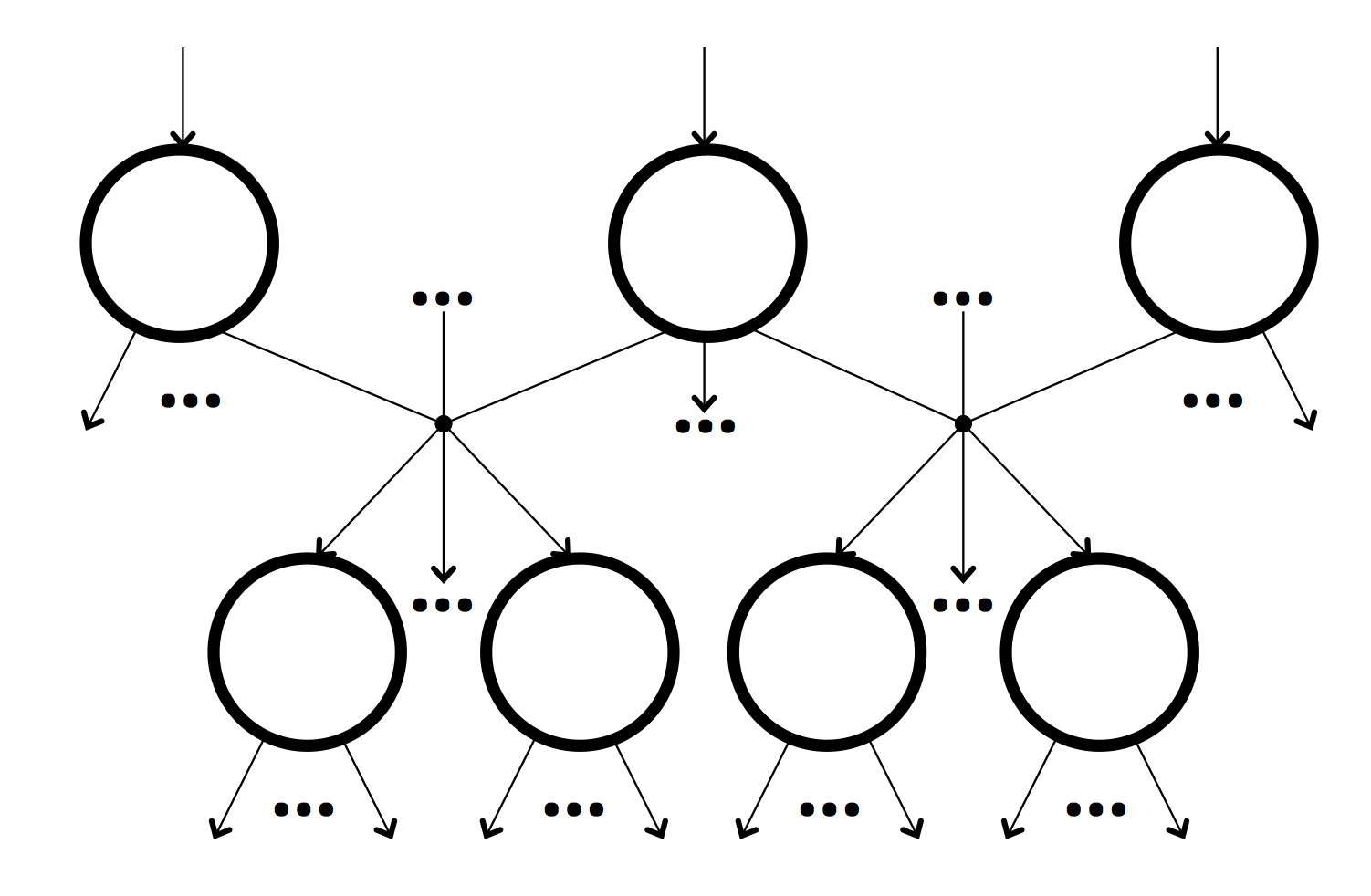}
    \caption{Hypergraph general form}
    \label{fig:hypergraph_general}
  \end{subfigure}
  \begin{subfigure}{0.3\linewidth}
    \includegraphics[height=1in]{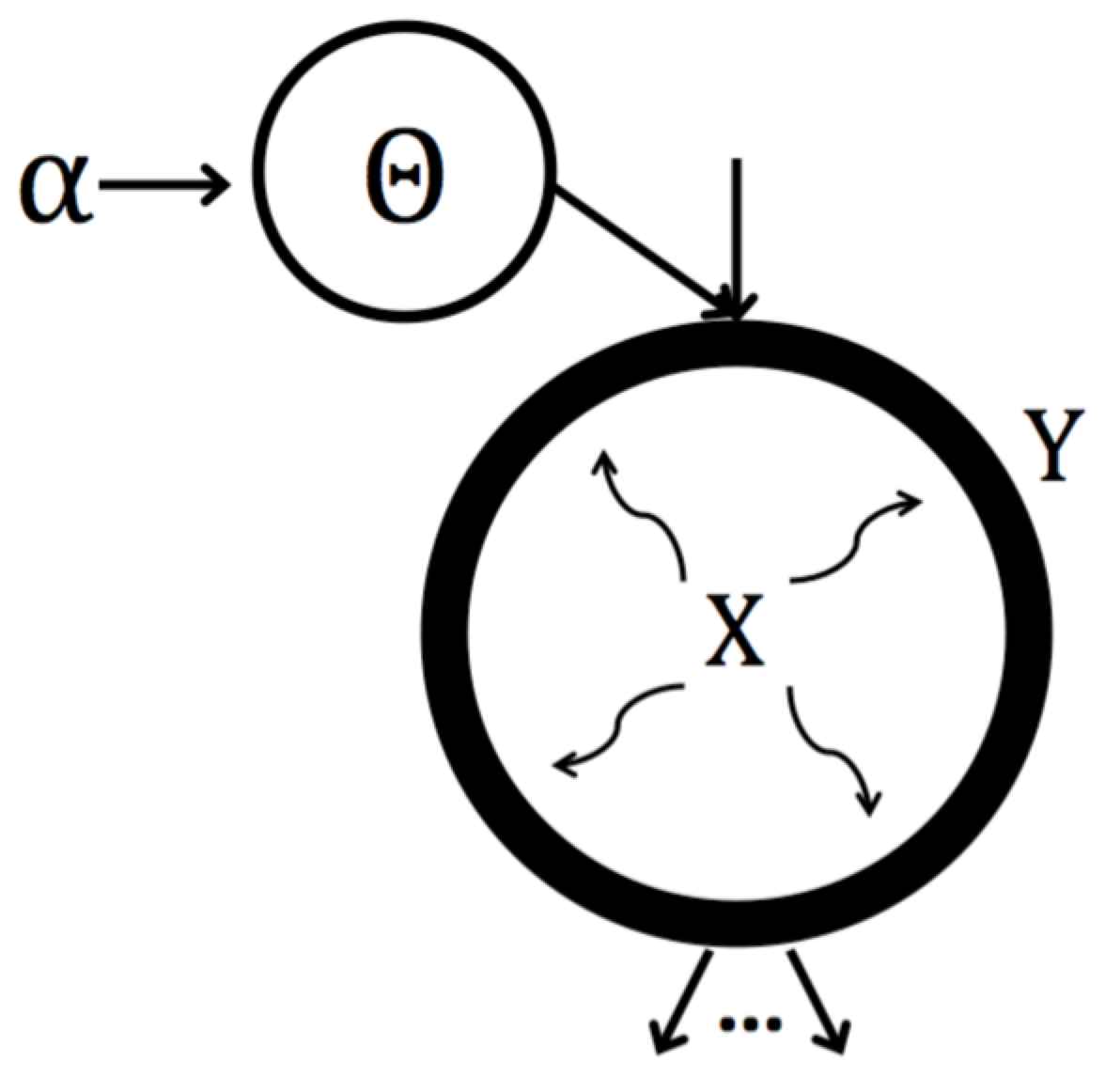}
    \caption{A node}
    \label{fig:node_detail}
  \end{subfigure}
  \caption{An example pedigree (a), is modeled by the general form of a hypergraph (b).  Latent ($X$) and emission ($Y$) states are determined by parents \newfix{and parameters ($\theta$)} (c).}
  \label{fig:hypergraph_examples}
\end{figure}

We represent pedigrees as multiply-connected directed acyclic hypergraphs.  A hypergraph is a generalization of a graph where edges connect more than two nodes (Figure \ref{fig:hypergraph_general}).  Multiply-connected directed acyclic hypergraphs have directed edges, contain no path out from a node that leads back to itself, and can have more than one directed path between any two nodes.  This structure sufficiently represents reproduction even when relatives have a child together.  We applied latent state space inference algorithms to this structure, described in Appendix \ref{appendix:polytree_algorithm}.

\subsection{Mendelian Inheritance as a Latent State Space Model}
Mendelian inheritance can be modeled using latent state space models.  Genotypes are latent states, phenotypes are emission states, and parameters are derived from Mendel's laws.  \textit{De novo} mutations are represented as unaffected to mutated state changes in the transition distribution and incomplete penetrance as mutated state to unaffected emission in the emission distribution.  In accordance with Mendelian genetics, we use discrete values for the latent and emission states.  We review the possible latent and emission states for AD, AR and XL in Appendix \ref{appendix:mendelian_priors}.

We placed Dirichlet priors over the rows of the root, transition and emission parameters and used the laws of Mendelian genetics to construct the expected transition and emission tensors.  The diseases we model are rare; thus we set the root priors to favor no genetic mutation.  There is a different set of parameters for each sex in the XL model to accommodate the different latent state types.  The full generative model is described in Appendix \ref{appendix:gen_model}.

We found that it was difficult to learn parameters due to our small and incomplete dataset.  Therefore, we avoided the learning problem by making predictions according to the rules of Mendelian inheritance, estimating the marginal probability under each inheritance pattern prior: $P(Y;\{\text{AD},\text{AR},\text{XL}\}) = E_{\theta \sim P(\theta;\{\text{AD},\text{AR},\text{XL}\})}[P(Y|\theta)]$.

\subsection{Hypothetical evidence}
Our implementation allows for \newfix{hypothetical evidence of the true} genotypes in a family tree.  \newfix{The approach we took was motivated by} the question "what if this person actually did inherit this genotype?" in the context of predicting inheritance patterns.  The intent is to allow humans to compare guesses of reality.  Practically, we \newfix{do this} by forcing specific state transitions (Appendix \ref{appendix:intervention}).  As a consequence, we can restrict the possible states while still accounting for the likelihood of the restriction.  By using \newfix{hypothetical evidence}, a specialist can directly see the implications that a hypothesis has on both the genotype probabilities for relatives and the inheritance pattern prediction.  To take advantage of the certainty of blood tests as it pertains the shading of a node, \newfix{we treated the blood test results as evidence that shaded nodes have carrier genotypes} (Appendix \ref{appendix:mendelian_priors}).  A human can further use intuition to guess the disease status of individuals in order to adjust their final prediction.

\begin{figure}
  \centering
  \begin{subfigure}[b]{0.35\linewidth}
    \includegraphics[width=\linewidth]{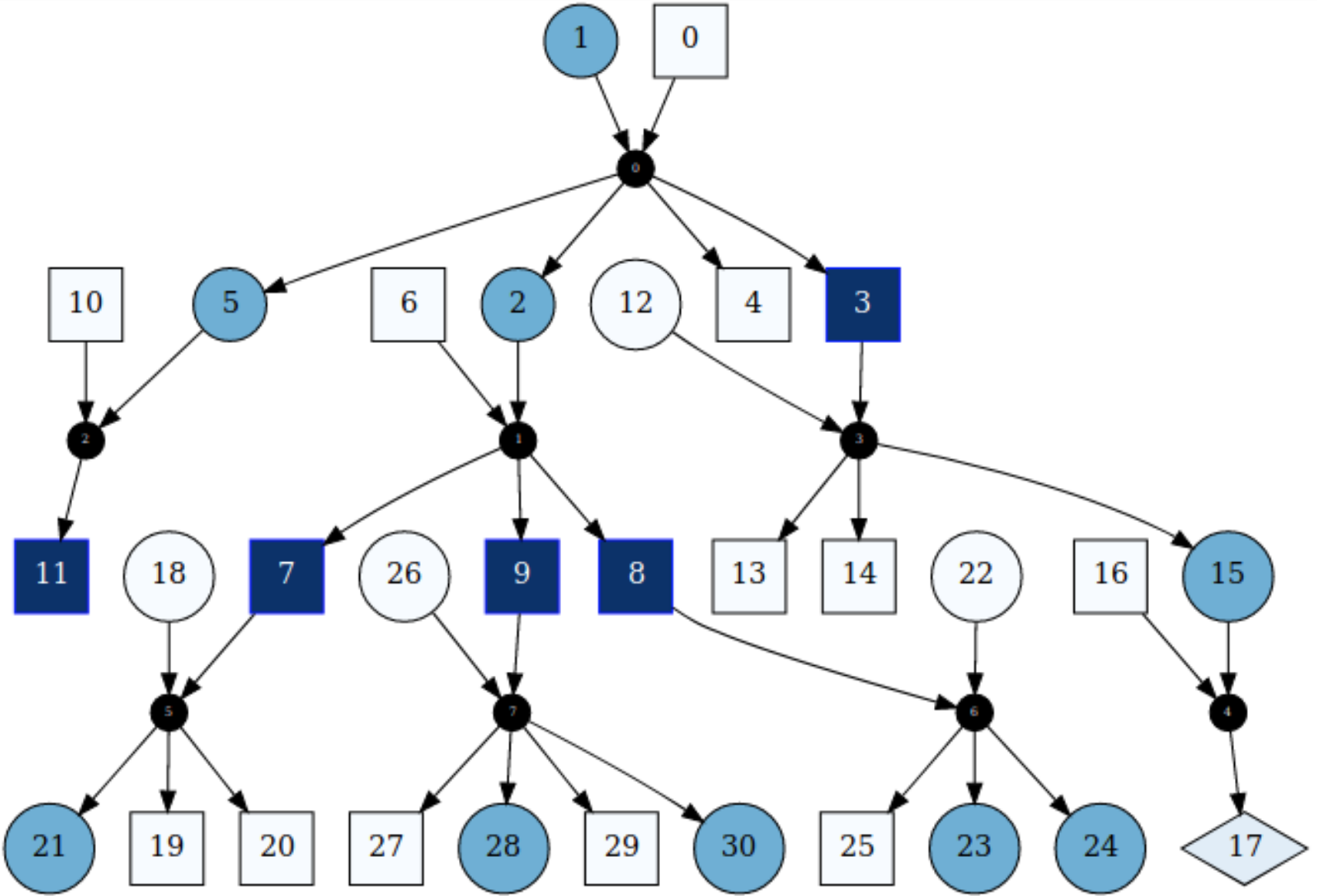}
    \caption{Inferred latent states on a pedigree}
    \label{fig:smoothed_example}
  \end{subfigure}
  \begin{subfigure}[b]{0.3\linewidth}
    \includegraphics[width=\linewidth]{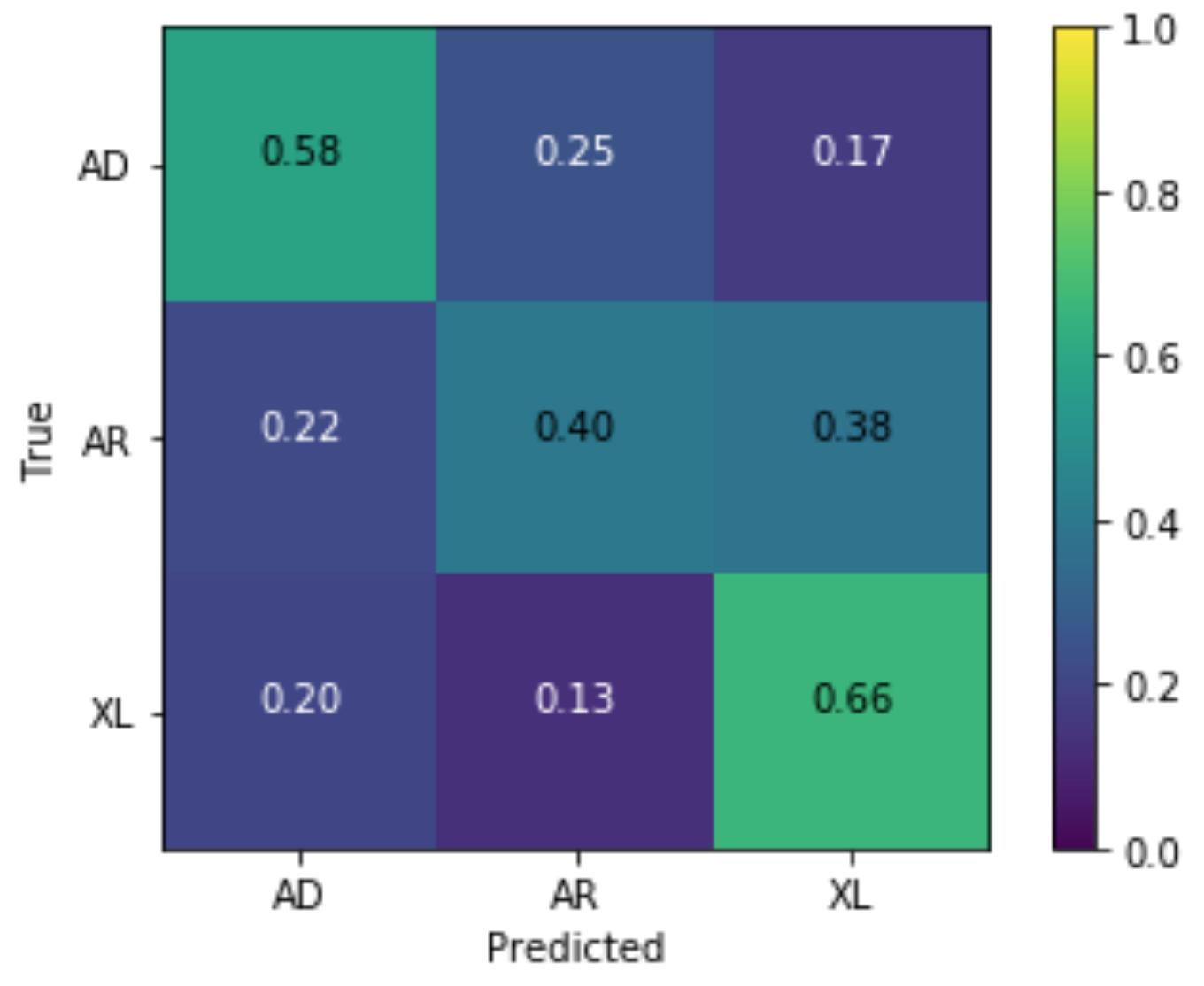}
    \caption{Accuracy on pedigrees}
    \label{fig:prediction_results_all}
  \end{subfigure}
  \begin{subfigure}[b]{0.3\linewidth}
    \includegraphics[width=\linewidth]{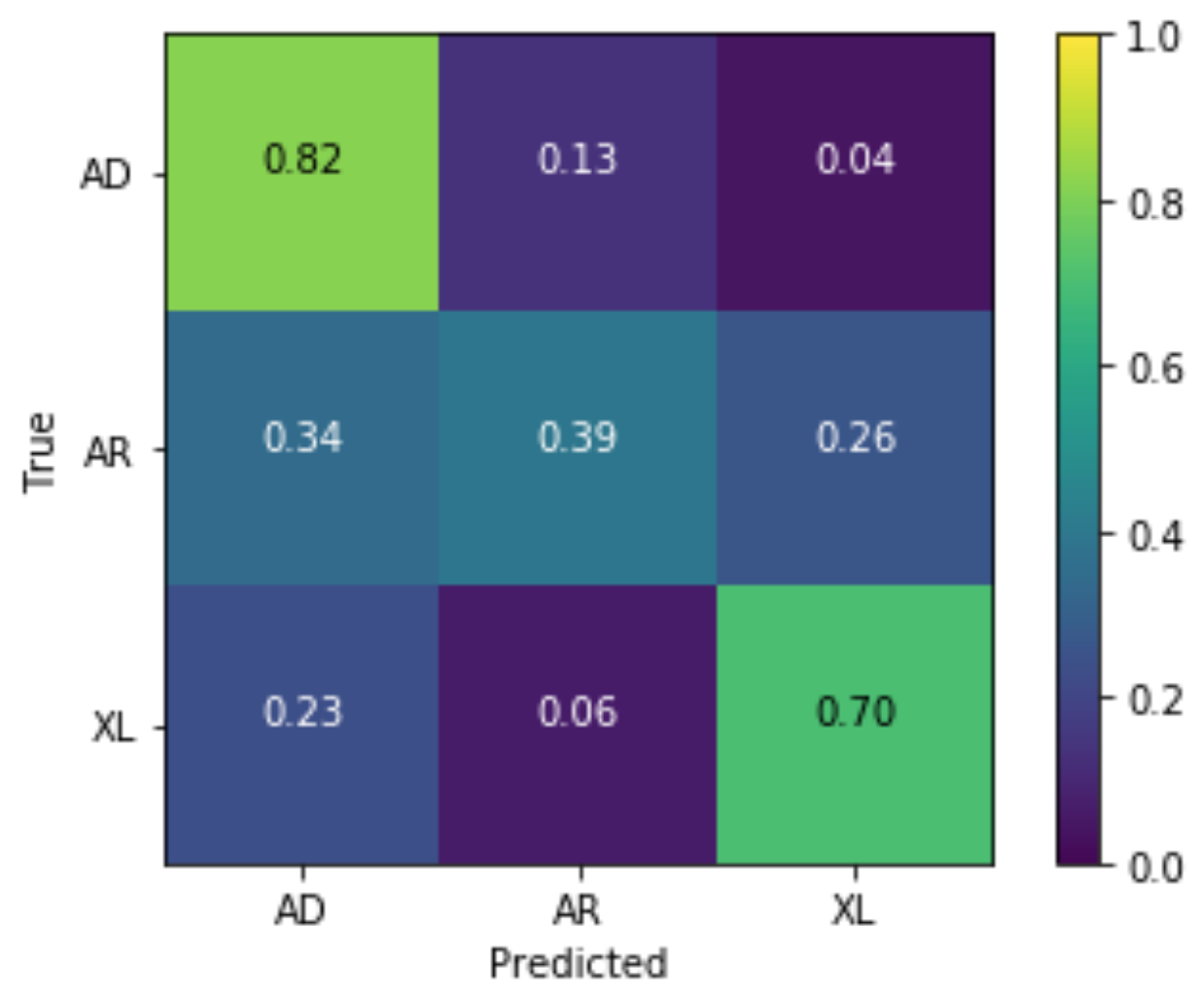}
    \caption{Only high confidence}
    \label{fig:prediction_results_confident}
  \end{subfigure}
  \caption{The smoothed latent states are depicted in (a).  We show the prediction accuracy of all pedigrees (N=427) (b) and those predicted with high confidence only (N=214) (c).}
  \label{fig:results}
\end{figure}

\subsection{Explainability}
Latent state space model inference algorithms make our approach explainable.  A direct consequence of making a prediction is access to a variety of probabilities associated with the likelihood of any node(s) actually having some genotype (Appendix \ref{appendix:smoothed_vals}).

One example is the distribution over possible alleles for any node.  A human can examine these values to see who is likely a carrier.  Another example is the distribution of the alleles conditioned on parent alleles.  A human can compare these values against what Mendelian inheritance expects to identify where \textit{de novo} mutations or incomplete penetrance likely occurred.  Because \textit{de novo} mutations are rare, predictions that expect many such occurrences might be ignored.

\section{Results}
We evaluated our algorithm on 132 AD, 197 AR, and 98 XL pedigrees.  We estimated the marginal probability of the data using 100 Monte Carlo samples from $E_{\theta \sim P(\theta;\alpha )}[P(Y|\theta)]$ for each inheritance pattern model and then chose the model with the largest value.  We closely followed Mendel's laws by using a prior strength of 1000000.  We selected this value by hand to mitigate the AR bias problem described in the Limitations (Section \ref{limitationssection}).  We left the optimization of hyperparameters for future work.  More details on the priors are in Appendix \ref{appendix:mendelian_priors}.  We delineated confident predictions as those with a max normalized marginal over each model greater than 80\%, mimicking the real-life situation where a geneticist cannot confidently determine an inheritance pattern for a particular pedigree.

\subsection{Explainability}
We are able to assign probabilities corresponding to genetics values that a human would only be able to estimate from intuition.  The example in Figure \ref{fig:smoothed_example} is the smoothed representation of Figure \ref{fig:pedigree_example}.  The color of each node represents how likely it is that the person has a mutated allele.  \newfix{The blood test result used to determine the shading of nodes was used as evidence to force males 3, 7, 8, 9 and 11 to have the mutated allele.}  The light shading indicates the algorithm's prediction that the affected males' mothers and daughters are expected to be carriers.  This agrees with human intuition on who may be a carrier.  The exact smoothed probabilities for each person are included in Appendix \ref{appendix:smooth_example}.

\subsection{Accuracy vs. Ground Truth}
We examined predictions on all pedigrees (Figure \ref{fig:prediction_results_all}) and on pedigrees that yielded confident predictions (Figure \ref{fig:prediction_results_confident}).  We were accurate 52\% of the time with a Cohen Kappa score of 0.29 over the entire dataset.  However, upon closer inspection by a genetic specialist, we found that many of the pedigrees in our dataset did not indicate a clear pattern of inheritance.

In real life, a geneticist will sometimes conclude that they cannot determine an inheritance pattern.  We mimicked this approach by counting only confident predictions, yielding an accuracy of 66\% with a Cohen's Kappa of 0.48 over 89 AD, 61 AR and 64 XL pedigrees.  The prediction accuracy on AD and XL of (77\%) was equal to that produced by \cite{Schlegel17}, who classified using high level features derived from both pedigree shading and extra annotations that our method did not use.  Pedigrees with only one shaded node are difficult to classify without more information.  139/197 of the AR pedigrees had only one shaded node, which could explain the low AR accuracy.

\subsection{Limitations}\label{limitationssection}

The rarity of the diseases in our dataset, and the inability to distinguish missing data from unaffected individuals hindered our method's performance.  Our initial analysis favored AR predictions because our dataset contained mostly unshaded nodes -- a feature that AR is most likely to produce.  To address this problem, we set the roots to favor no genetic mutation and also strengthened the transition distribution to lower the \textit{de novo} mutation rate.  The strength values were selected by hand and therefore may not reflect the true rates for the diseases.  The largest limitation was that we could not distinguish missing data from unaffected people.  As a result, our prediction incorporated more instances of incomplete penetrance than it would have otherwise.

\section{Related Work}
Genetic specialists use high level features to infer the most likely inheritance pattern of a disease.  Their approach works well because the features are a consequence of Mendelian inheritance, and the genetic specialist can make judgments about what parts of the pedigree are important.  \cite{Schlegel17} used features engineered by genetic specialists to construct a gradient boosting tree model to predict the inheritance pattern.  Although this method achieved reasonable accuracy, its results were not explainable.

Bayesian networks provide generative models for arbitrary graphs, as described by \cite{Pearl85}.  \cite{Pearl86}'s belief propagation algorithm allows for message passing for polytrees while the junction tree algorithm \cite{Lauritzen90} and cutset conditioning \cite{Pearl86} provide inference over general graphs, \newfix{but we used a custom message passing algorithm (Appendix \ref{appendix:polytree_algorithm}) to provide inference over the latent states.}  \cite{Sun2007HaplotypeIU} modeled genetic inheritance using a Higher-Order Hidden Markov Model for haplotype inference but did not examine different types of genetic inheritance.

\section{Conclusions}
Inheritance patterns are important to the diagnosis of an inherited disease.  We modeled inheritance patterns as a latent state space model in order to provide explainable predictions.  We believe that our work can help genetic specialists provide patient care in cases of rare inherited diseases.

\newpage
\bibliographystyle{ACM-Reference-Format}
\bibliography{bibliography}

\newpage
\section*{Appendix}

\subsection{Mendelian Priors} \label{appendix:mendelian_priors}
We derived the priors for the transition parameters from Punnet squares.  In the AD and AR models, the possible latent states are $AA$, $Aa$ or $aa$, and in the XL model the possible latent states are $X^AX^A$, $X^AX^a$ or $X^aX^a$ for females, $X^AY$, $X^aY$ for males and $X^AX^A$, $X^AX^a$, $X^aX^a$, $X^AY$, $X^aY$ for unknown sex.  The emission states correspond to the shading of each node on the pedigree, which denote diagnosed individuals.

Below is a full autosomal Punnet square.  Each axis represents a parent and the grid represents the combined child autosome pair.  During reproduction using Mendelian inheritance, we would select a child pair of alleles given its parents' alleles using this Punnet square - the intersection of a row and a column of the square yields a 2x2 block that represents the possible child latent states.  We distinguish the Aa and aA states here for clarity, but we combined them into one state in our model.

\[
\begin{tabular}[t]{r|r|c|c|c|c|c|c|c|c|}
\multicolumn{1}{r}{}
&\multicolumn{1}{r}{}
&\multicolumn{1}{c}{AA}
&\multicolumn{1}{c}{}
&\multicolumn{1}{c}{Aa}
&\multicolumn{1}{c}{}
&\multicolumn{1}{c}{aA}
&\multicolumn{1}{c}{}
&\multicolumn{1}{c}{aa}
&\multicolumn{1}{c}{}
\\\cline{2-10}
 &&A&A&A&a&a&A&a&a\\\cline{2-10}
AA&A&AA&AA&AA&Aa&Aa&AA&Aa&Aa\\\cline{3-10}
  &A&AA&AA&AA&Aa&Aa&AA&Aa&Aa\\\cline{2-10}
Aa&A&AA&AA&AA&Aa&Aa&AA&Aa&Aa\\\cline{3-10}
  &a&aA&aA&aA&aa&aa&aA&aa&aa\\\cline{2-10}
aA&a&aA&aA&aA&aa&aa&aA&aa&aa\\\cline{3-10}
  &A&AA&AA&AA&Aa&Aa&AA&Aa&Aa\\\cline{2-10}
aa&a&aA&aA&aA&aa&aa&aA&aa&aa\\\cline{3-10}
  &a&aA&aA&aA&aa&aa&aA&aa&aa\\\cline{2-10}
\end{tabular}
\]
\newline
\newline
The final transition prior is indexed using the mother, father and child states to index into the first, second and third axis of the tensor respectively.
\[
\alpha^\text{AD and AR transition} :=
\begin{blockarray}{[c[ccc]c[ccc]c[ccc]c]}
    & 1   & 0   & 0 &   & 0.5  & 0.5 & 0    &   & 0   & 1   & 0    & \\
    & 0.5 & 0.5 & 0 &   & 0.25 & 0.5 & 0.25 &   & 0   & 0.5 & 0.5  & \\
    & 0   & 1   & 0 & , & 0    & 0.5 & 0.5  & , & 0   & 0   & 1    & \\
\end{blockarray}
^T
\]

Using the same approach for sex in X-linked inheritance, we can compute the transition priors for each sex:
\[
\alpha^\text{XL female transition} :=
\begin{blockarray}{[c[ccc]c[ccc]c[ccc]c]}
    & 1   & 0   & 0 &   &  0.5  & 0.5 & 0   &   & 0   & 1   & 0 & \\
    & 0   & 1   & 0 & , &  0    & 0.5 & 0.5 & , & 0   & 0   & 1 & \\
\end{blockarray}
^T
\]

\[
\alpha^\text{XL male transition} :=
\begin{blockarray}{[c[cc]c[cc]c[cc]c]}
    & 1   & 0 &   & 0.5  & 0.5 &   & 0 & 1 & \\
    & 0   & 1 & , & 0.5  & 0.5 & , & 0 & 1 & \\
\end{blockarray}
^T
\]

\[
\alpha^\text{XL unknown sex transition} :=
\left[
\begin{blockarray}{ccccc}
\begin{block}{[ccccc]}
    0.5   & 0     & 0 & 0.5   & 0 \\
    0     & 0.5   & 0 & 0     & 0.5 \\
\end{block}
\begin{block}{[ccccc]}
    0.25  & 0.25 & 0    & 0.25  & 0.25 \\
    0     & 0.25 & 0.25 & 0.25  & 0.25  \\
\end{block}
\begin{block}{[ccccc]}
    0   & 0.5   & 0   & 0   & 0.5 \\
    0   & 0     & 0.5 & 0   & 0.5 \\
\end{block}
\end{blockarray}
\right]
\]

The emission priors come directly from the definition of dominant and recessive.  These values correspond to the probability of a node being shaded on a pedigree given a certain latent state.  Under Mendelian inheritance, the states that produce a shaded node are ($AA,Aa$) in AD, ($aa$) in AR, ($X^aX^a$) for females in XL, ($X^aY$) for males in XL and ($X^aX^a$, $X^aY$) for unknown sex in XL.  We set the last index of the rows of each matrix to denote the affected state.
\[
\alpha^\text{AD emission} :=
\begin{bmatrix}
    0 & 1 \\
    0 & 1 \\
    1 & 0
\end{bmatrix}
\]
\[
\alpha^\text{AR emission} :=
\begin{bmatrix}
    0 & 1 \\
    1 & 0 \\
    1 & 0
\end{bmatrix}
\]
\[
\alpha^\text{XL female emission} :=
\begin{bmatrix}
    0 & 1 \\
    1 & 0 \\
    1 & 0
\end{bmatrix}
\]
\[
\alpha^\text{XL male emission} :=
\begin{bmatrix}
    0 & 1 \\
    1 & 0
\end{bmatrix}
\]
\[
\alpha^\text{XL unknown sex emission} :=
\begin{bmatrix}
    0 & 1 \\
    1 & 0 \\
    1 & 0 \\
    0 & 1 \\
    1 & 0
\end{bmatrix}
\]

The rows of the transition and emissions values we've defined are priors of a Dirichlet distribution over rows of the transition and emission parameters respectively.  Dirichlet random variables are typically used in Bayesian inference as priors over Categorical or Multinomial distributions.  The probability density function of a random variable $\theta \sim \text{Dirichlet}(\alpha)$ is:
\[
\frac{\Gamma(\sum_{i=1}^K\alpha_i)}{\prod_{i=1}^K\Gamma(\alpha_i)}\prod_{i=1}^K(\theta_i^{\alpha_i-1})
\]
where $\alpha$ is a $K$-vector and $\Gamma(x)$ is the Gamma function.  Random variables $\theta$ lie in the $K-1$ simplex ($\theta_i \geq 0, \sum_{i=1}^K\theta_i=0$) and are therefore valid Categorical distributions.

The $\alpha$ parameter can be thought of as the expected class counts, $(c_1,\dots,c_K)$, that would be simulated from $X \sim P(X|\theta;\alpha)$.  The larger the values of $\alpha$, the closer $(c_1,\dots,c_K)$ will be to $\alpha$.

For Mendelian inheritance, this implies that larger values of $\alpha$ will yield parameters that follow the three laws more closely.  Conversely, smaller values increase the \textit{de-novo} mutation rate and incomplete penetrance rate.  To control the deviation from strict Mendelian inheritance, we used a prior-strength hyperparameter that calculated the final prior that was used in the model as:
\[
\alpha^{Final} = 1 + \alpha*(\text{Prior Strength})
\]

\subsection{Smoothing example} \label{appendix:smooth_example}
This is the full list of smoothed probabilities corresponding to figure \ref{fig:smoothed_example}.

\begin{blockarray}{c|cc||c|ccc||c|ccccc}
Male & XY & xY & Female & XX & Xx & xx & Unknown & XX &  Xx   & xx   & XY   & xY   \\
\begin{block}{c|cc||c|ccc||c|ccccc}
   0 & 0. & 1. &      1 & 0. & 1. & 0. &      17 & 0. &  0.33 & 0.33 & 0.   & 0.33 \\
\end{block}
\begin{block}{c|cc||c|ccccccccc}
   3 & 1. & 0. &      2 & 0. & 1. & 0. &         &    &       &      &      &      \\
   4 & 0. & 1. &      5 & 0. & 1. & 0. &         &    &       &      &      &      \\
   6 & 0. & 1. &     12 & 0. & 0. & 1. &         &    &       &      &      &      \\
   7 & 1. & 0. &     15 & 0. & 1. & 0. &         &    &       &      &      &      \\
   8 & 1. & 0. &     18 & 0. & 0. & 1. &         &    &       &      &      &      \\
   9 & 1. & 0. &     21 & 0. & 1. & 0. &         &    &       &      &      &      \\
  10 & 0. & 1. &     22 & 0. & 0. & 1. &         &    &       &      &      &      \\
  11 & 1. & 0. &     23 & 0. & 1. & 0. &         &    &       &      &      &      \\
  13 & 0. & 1. &     24 & 0. & 1. & 0. &         &    &       &      &      &      \\
  14 & 0. & 1. &     26 & 0. & 0. & 1. &         &    &       &      &      &      \\
  16 & 0. & 1. &     28 & 0. & 1. & 0. &         &    &       &      &      &      \\
  19 & 0. & 1. &     30 & 0. & 1. & 0. &         &    &       &      &      &      \\
\end{block}
\begin{block}{c|cccccccccccc}
  20 & 0. & 1. &        &    &    &    &         &    &       &      &      &      \\
  25 & 0. & 1. &        &    &    &    &         &    &       &      &      &      \\
  27 & 0. & 1. &        &    &    &    &         &    &       &      &      &      \\
  29 & 0. & 1. &        &    &    &    &         &    &       &      &      &      \\
\end{blockarray}

\subsection{Generative Model Description}
\label{appendix:gen_model}
For each inheritance pattern, the generative model is as follows:
Sample parameters from the Mendelian priors:

\begin{align*}
\pi_0 &\sim \text{Dirichlet}( \alpha^\text{root} ), \\
\pi_{ij} &\sim \text{Dirichlet}( \alpha^\text{transition}_{ij} ), \\
L_i &\sim \text{Dirichlet}( \alpha^\text{emission}_{i} )
\end{align*}

For each root:
\begin{align*}
x^r &\sim \text{Categorical}( \pi_0 ), \\
y^r &\sim \text{Categorical}( L_{x^r} )
\end{align*}

Then perform a breadth first search of the graph.  For each child $c$ with mother $m$ and father $f$:
\begin{align*}
x^c &\sim \text{Categorical}( \pi_{x^m x^f} ), \\
y^c &\sim \text{Categorical}( L_{x^c} )
\end{align*}

\subsection{Evidence} \label{appendix:intervention}
Our algorithm made use of \newfix{hypothetical evidence in the} latent states to test guesses about the true latent states in the system.  \newfix{We denoted evidence for state $n_x$ as $S_n$.}  For all nodes $N$ and roots $R$, the joint distribution $P(N_x,N_y,\theta|\doOp{N_x \in S_N})$ of our model with \newfix{evidence} can be shown to factor to:
\[
P(\theta;\alpha)\prod_{r \in R}P(r_x|\theta,\doOp{r_x \in S_r})\prod_{n \in N \setminus R}P(n_x|pa(n)_x,\doOp{n_x \in S_n},\theta)\prod_{n \in N}P(n_y|n_x,\theta)
\]
We computed $P(r_x|\theta,\doOp{r_x \in S_r})$ and $P(n_x|pa(n)_x,\doOp{n_x \in S_n},\theta)$ using:
\begin{align*}
P(r_x|\theta,\doOp{r_x \in S_r}) &= P(r_x|\theta)*I[r_x \in S_r] \\
P(n_x|pa(n)_x,\doOp{n_x \in S_n},\theta) &= P(n_x|pa(n)_x,\theta)*I[n_x \in S_n]
\end{align*}
where I[.] is an indicator function of the appropriate shape.

\newfix{Providing hypothetical evidence for} latent states using this approach has two consequences.  The first is the same Pearl's do operator -- inference is run over the entire graph under the assumption that there are restrictions on the latent states.  The second is that predictions take into account how likely the \newfix{guesses} are.  This is crucial for comparing its implications across the different inheritance patterns.

In our application, \newfix{hypothetical evidence is} meant to be \newfix{a guess} of reality.  Bad guesses that disagree with Mendel's laws should not have the same weight as good guesses.  Without accounting for the likelihood of the \newfix{guess}, it would be trivial to get a large marginal value for any inheritance pattern model.

\subsection{Inference algorithm derivation}\label{appendix:polytree_algorithm}
We use the term ``up edge'' to refer to the edge for which a node $n$ is in its child set, and ``down edges'' for the edges that $n$ is a parent of.  Nodes that are also children of $n$'s up edge are $n$'s siblings.  Nodes that are a part of a parent set with $n$ are called the mates of $n$, and the nodes in the child set of an edge of which $n$ is a parent of are called the children of $n$.  The up edge, down edges and set of parents and siblings are denoted as $\wedge(n)$, $\vee(n)$, $P(n)$ and $S(n)$, respectively. For a down edge $e$ of $n$, the mates and children are denoted by $M(n,e)$, $C(n,e)$ respectively.  For a node $n$, $\uparrow(n)$ denotes all nodes that can be reached by branching up $n$'s up edge, $\downarrow(n,e)$ denotes all nodes that can be reached by branching down $n$'s down edge $e$, and $!(n,e)$ denotes all nodes that can be reached without branching down $e$ from $n$.  Nodes with no up edge are called the roots (the oldest ancestor in a pedigree) and nodes with no down edges are called the leaves (the youngest ancestor in a pedigree).  Finally, each node has a type, which we use to model biological sex.

Let $Y$ be the set of all emission states, $a(n,e) := \mya{n}{e}$, $b(n) := \myb{n}$, $u(n) := \myu{n}$ and $v(n,e) := \myv{n}{e}$.
The inference algorithm we used over polytrees is as follows:

For roots $r$ and leaves $l$ of $G$:
\begin{align*}
u(r) &= P(r_y|r_x)P(r_x) \\
v(l) &= 1
\end{align*}

For all other nodes:
\begin{align*}
    a(n,e) &= u(n)\downedgeprodexcept{n}{e}v(n,e\prime) \\
    b(n) &= \int_{n_x}\transition{n}\downedgeprod{n}v(n,e)dn_x \\
    u(n) &= \int_{P(n)_x}\transition{n}\siblingprod{n}b(n_s)\parentprod{n}a(n_p,\wedge(n))dP(n)_x \\
    v(n,e) &= \int_{M(n,e)_x}\childprod{n}{e}b(n_c)\mateprod{n}{e}a(n_m,e)dM(n,e)_x \\
    P(n_x,Y) &= u(n)\prod_{e\in \vee(n)}v(n,e), \text{          } P(Y) = \int_{n_x}P(n_x,Y)dn_x
\end{align*}

Below, we derive each of these updates

\begin{align*}
    a(n,e) &= \mya{n}{e} \\
    b(n)   &= \myb{n} \\
    u(n)   &= \myu{n} \\
    v(n,e) &= \myv{n}{e}
\end{align*}

\subsubsection{a derivation}
Intuitively, $a(n,e)$ is the probability of all emission states that can be reached without traversing $e$ from $n$ and $n$'s latent state.
\begin{align*}
    a(n,e) &= \mya{n}{e} \\
           &= P( n_y, \uparrow(n)_y, \downedgeandexcept{n}{e}\downarrow(n,e\prime)_y, n_x ) \\
           &= \myu{n}P(\downedgeandexcept{n}{e}\downarrow(n,e\prime)_y|n_x) \\
           &= \myu{n}\downedgeprodexcept{n}{e}\myv{n}{e
           \prime} \\
           &= u(n)\downedgeprodexcept{n}{e}v(n,e\prime)
\end{align*}

\subsubsection{b derivation}
Intuitively, $b(n)$ is the probability of all emission states (including $n$'s) that can be reached without traversing $n$'s up edge, given the latent states of the parents of $n$.
\begin{align*}
    b(n) &= \myb{n} \\
         &= \int_{n_x}P(\downarrow(n)_y,n_y,n_x|P(n)_x)dn_x \\
         &= \int_{n_x}P(\downarrow(n)_y,n_y|n_x)P(n_x|P(n)_x)dn_x \\
         &= \int_{n_x}P(\downedgeand{n}\downarrow(n,e)_y,n_y|n_x)P(n_x|P(n)_x)dn_x \\
         &= \int_{n_x}\transition{n}\downedgeprod{n}\myv{n}{e}dn_x \\
         &= \int_{n_x}\transition{n}\downedgeprod{n}v(n,e)dn_x
\end{align*}

\subsubsection{u derivation}
Intuitively, $u(n)$ is the probability of all emission states (including $n$'s) that can be reached by traversing $n$'s up edge and $n$'s latent state.
\begin{align*}
    \begin{split}
    u(n) {}&= \myu{n}
    \end{split} \\
    \begin{split}
         {}&= P(\parentand{n}!(n_p,\wedge(n))_y, \siblingand{n}(n_{s_y},\downarrow(n_s)_y),n_y,n_x)
    \end{split} \\
    \begin{split}
         {}&= \int_{P(n)_x}P(\parentand{n}!(n_p,\wedge(n))_y, \siblingand{n}(n_{s_y},\downarrow(n_s)_y),n_y,n_x,P(n)_x)dP(n)_x
    \end{split} \\
    \begin{split}
         {}&= \int_{P(n)_x}P(\siblingand{n}(n_{s_y},\downarrow(n_s)_y),n_y,n_x|P(n)_x) P(\parentand{n}!(n_p,\wedge(n))_y,P(n)_x) dP(n)_x
    \end{split} \\
    \begin{split}
         {}&= \int_{P(n)_x}\transition{n}\siblingprod{n}\myb{n_s} \\
           &\parentprod{n}\mya{n_p}{\wedge(n)} dP(n)_x
    \end{split} \\
         {}&= \int_{P(n)_x}\transition{n}\siblingprod{n}b(n_s)\parentprod{n}a(n_p,\wedge(n))dP(n)_x
\end{align*}

\subsubsection{v derivation}
Intuitively, $v(n,e)$ is the probability of all emission states that can be reached by traversing $e$ from $n$ given $n$'s latent state.
\begin{align*}
    v(n,e) &= \myv{n}{e} \\
           &= P(\mateand{n}{e}!(n_m,e)_y, \childand{n}{e}(n_{c_y},\downarrow(n_c)_y)|n_x) \\
           &= \int_{M(n,e)_x}P(\mateand{n}{e}!(n_m,e)_y, \childand{n}{e}(n_{c_y},\downarrow(n_c)_y),M(n,e)_x|n_x)dM(n,e)_x \\
           &= \int_{M(n,e)_x}P(\childand{n}{e}(n_{c_y},\downarrow(n_c)_y)|M(n,e)_x,n_x)P(\mateand{n}{e}!(n_m,e)_y,M(n,e)_x|n_x)dM(n,e)_x \\
           &= \int_{M(n,e)_x}\childprod{n}{e}\myb{n_c}\mateprod{n}{e}\mya{n_m}{e}dM(n,e)_x \\
           &= \int_{M(n,e)_x}\childprod{n}{e}b(n_c)\mateprod{n}{e}a(n_m,e)dM(n,e)_x
\end{align*}

For multiply-connected graphs, we slightly modify the algorithm.  For nodes $N$ and edges $G$, let $G = <N,E>$ be the original graph.  The general algorithm first finds a feedback vertex set $F \subset N$ of $G$.  $F$ is defined to be a set that $G_F = <N-F,E>$ is a polytree with one connected component.  Then the polytree algorithm is run on $G_F$ using the following recursive equations:

For roots $r$ and leaves $l$ of $G$:
\begin{align*}
u_F(r) &= P(r_y|r_x)P(r_x) \\
v_F(l) &= 1
\end{align*}

For the rest of the nodes, let $P_F(n)=P(n)\setminus F$ and $M_F(n)=M(n)\setminus F$
\begin{align*}
a_F(n,e) &= u_F(n)\downedgeprod{n}v_F(n,e) \\
&\text{If }n\in F \\
b_F(n)   &= \int_{n_x}\transition{n}\downedgeprod{n}v_F(n,e)dn_x \\
&\text{Else} \\
b_F(n)   &= \transition{n} \\
u_F(n)   &= \int_{P_F(n)_x}\transition{n}\siblingprod{n}b_F(n_s)\parentFprod{n}a_F(n_p,\wedge(n))dP_F(n)_x \\
v_F(n,e) &= \int_{M_F(n,e)_x}\childprod{n}{e}b_F(n_c)\mateFprod{n}{e}a_F(n_m,e)dM_F(n,e)_x
\end{align*}

Using these updates, we were able to achieve exact inference over multiply-connected hypergraphs with computational complexity that is exponential with the number of nodes in the feedback vertex set.

\subsection{Computing Smoothed Values}\label{appendix:smoothed_vals}

Using the values computed above, we can compute a variety of probabilities associated with the latent states, conditioned on the emissions.

For nodes $n \notin F$:
\begin{align*}
P(n_x,Y) &= \int_{F_x}u_F(n)\downedgeprod{n}v_F(n,e) dF_x \\
         &= \int_{F_x}P(n_x,F_x,Y)dF_x
\end{align*}
For feedback node $f \in F$ and any node $n \notin F$, let $F_{f,n}$ denote $F \setminus \{f\} \setor \{n\}$:
\begin{align*}
P(f_x,Y) &= \int_{F_{{f,n}_x}}P(n_x,F_x,Y) dF_{{f,n}_x}
\end{align*}

At any node $n \in N$:

\begin{align*}
P(Y) &= \int_{n_x}P(n_x,Y)dn_x
\end{align*}

and

\begin{align*}
P(n_x|Y) &= P(n_x,Y)/P(Y)
\end{align*}

For any non-root $n \notin F$:

\begin{align*}
P(n_x,P(n)_x,Y) &= \int_{F_x}\transition{n}\siblingprod{n}b_F(n_s)\parentprod{n}a_F(n_p,\wedge(n))\downedgeprod{n}v_F(n,e) dF_x \\
                &= \int_{F_x}P(n_x,P(n)_x,F_x,Y) dF_x
\end{align*}

We can compute special cases where $n \in F$ or some or all of $n's$ parents are in $F$ by using the same trick used to calculate $P(f_x,Y)$ or by appropriately selecting nodes to integrate out of $P(n_x,P(n)_x,F_x,Y)$.

Using $P(n_x,P(n)_x,Y)$, we can trivially compute the following:

\begin{align*}
P(P(n)_x,Y) &= \int_{n_x}P(n_x,P(n)_x,Y)dn_x \\
P(n_x|P(n)_x,Y) &= P(n_x,P(n)_x,Y) / P(P(n)_x,Y)
\end{align*}

To verify the correctness of our algorithm, we ensured that every value above was a valid probability distribution ($P(.|Y)$ integrates to 1) and all values were consistent everywhere in the graph ($P(.,Y)$ was the same regardless of how/where it was computed).

\end{document}